\pdfoutput=1
\documentclass[mlmain]{jmlr}

\usepackage{tikz}
\usepackage{capt-of}
\usetikzlibrary{calc,patterns}
\definecolor{barfill}{RGB}{213,216,255}
\definecolor{barstroke}{RGB}{74,85,255}
\definecolor{heatnegstrong}{RGB}{54,76,100}
\definecolor{heatneg}{RGB}{139,164,184}
\definecolor{heatzero}{RGB}{248,246,241}
\definecolor{heatposlight}{RGB}{239,194,166}
\definecolor{heatpos}{RGB}{215,111,77}
\definecolor{heatposstrong}{RGB}{166,61,47}
\definecolor{heatmissing}{RGB}{245,243,238}

\jmlrvolume{}
\firstpageno{1}
\editors{}
\jmlryear{2026}
\jmlrworkshop{Symmetry and Geometry in Neural Representations}

\title[Operation-Selective Transfer Across Low-Overlap Circuits]
{Artificial Unintelligence? Operation Mismatch Limits Circuits' Use of Shared Representational Structure}

\author[Desai and Pendharkar]{\Name{Rishaan Desai} \Email{rishaandesai@gmail.com}\\
\addr BASIS Independent Silicon Valley
\AND
\Name{Ayan Pendharkar} \Email{ayan.pendharkar@gmail.com}\\
\addr Independent
}

\hypersetup{
  pdftitle={Artificial Unintelligence? Operation Mismatch Limits Circuits' Use of Shared Representational Structure},
  pdfauthor={Rishaan Desai and Ayan Pendharkar},
  pdfkeywords={mechanistic interpretability, representation geometry, activation patching, causal circuits, language models}
}

\begin{document}

\maketitle

\begin{abstract}
Language models often encode ordered concepts such as weekdays and months in low-dimensional circular subspaces. The causal role of this geometry is less clear: circuits for different tasks may organize their representations with analogous circular geometry without being able to use one another's computations. We test this possibility by discovering sparse task circuits and transplanting their activations across nine cyclic task families. We found that Qwen produced held-out circular geometry and selection-sufficient circuits for seven families; the resulting masks had little neuron overlap (median Jaccard $0.122$). Across 42 directed task pairs, a donor performing the target's signed shift increased normalized restoration by $0.651$ over a different-shift donor (95\% bootstrap CI $[0.474,0.838]$; sign-flip $p<10^{-6}$). The discrete operation match predicted repair, whereas continuous similarity between fitted operators was nearly uncorrelated with control-adjusted transfer strength ($\rho=0.031$). Activations from low-overlap task circuits can therefore support cross-task repair across circularly structured representations when their signed operations agree. A clean latent shape leaves that compatibility unresolved, so geometric claims about computation require operation-specific interventions.
\end{abstract}


\section{Introduction}
\label{sec:introduction}

Ask a language model what day falls twelve days after Wednesday. A successful
answer requires the model to locate Wednesday in a seven-element cycle, advance
twelve steps, and wrap around the cycle. The same abstract computation appears
in month arithmetic, clock positions, compass directions, alphabet shifts, and
musical notes. Internal activations for such concepts frequently occupy
low-dimensional geometric structures, including circles and other
manifolds~\citep{nanda2023progressmeasuresgrokkingmechanistic,gurnee2024languagemodelsrepresentspace,modell2025originsrepresentationmanifoldslarge}. This
geometry provides a compact account of where concepts are represented. It does
not by itself identify the mechanism that moves between them.

Mechanistic interpretability usually searches for a circuit: a set of model
components whose interventions explain a behavior
\citep{elhage2021mathematical,wang2022interpretabilitywildcircuitindirect,conmy2023automatedcircuitdiscoverymechanistic}. Prior work has found
substantial component reuse across related tasks~\citep{merullo2024circuitcomponentreusetasks}, and
recent work reports a shared set of MLP neurons that implements base-10
addition across several cyclic tasks in Llama~\citep{feucht2026arithmeticwildllamauses}.
Those results motivate a stricter question. If two task-specific circuits share
few neurons but act on representations with the same topology, can one circuit
supply a computation that the other task can use?

We answer this question with a donor-restoration experiment. For each cyclic
task, we fit a held-out circular subspace, discover a sparse MLP circuit, and
fit the affine operator induced by that circuit in the circular coordinates.
We then ablate the donor circuit while running target-task prompts and patch in
donor activations. The critical comparison holds the intervention pipeline
fixed and changes the donor's signed shift. If circular structure provides a
usable interface, a donor performing the target's shift should repair more of
the lost logit margin than a donor performing a different shift.

In Qwen, seven task families pass the behavioral, geometric, and circuit
gates. Their circuits overlap little, yet matched operations repair targets
substantially better than mismatched operations. Finer variation in our
continuous operator-similarity score does not explain which task pairs transfer
most strongly. The evidence supports operation-selective compatibility across
these task circuits; it does not establish that geometrically similar circuits
are generally functionally equivalent.

\section{Related Work}
\label{sec:related}

Fourier analyses of modular arithmetic have shown that networks can encode
periodic variables on circles and implement addition through rotations
\citep{nanda2023progressmeasuresgrokkingmechanistic}. Language-model activations also encode spatial and
temporal variables in linear subspaces~\citep{gurnee2024languagemodelsrepresentspace}, while
\citet{modell2025originsrepresentationmanifoldslarge} develop a broader account of how continuous features
can form representation manifolds. These papers explain how concept structure
appears in activation space. Our study examines the next causal step: whether
separately discovered mechanisms can exchange transformations through that
structure.

Cyclic concepts have also been studied at the component level. Successor heads
increment ordered tokens such as weekdays, months, letters, and numbers across
several model families~\citep{gould2023successorheadsrecurringinterpretable}. Their recurring behavior
suggests that related sequences can recruit task-general machinery. More
recently, \citet{feucht2026arithmeticwildllamauses} found that Llama-3.1-8B maps several
cyclic tasks through a shared base-10 addition mechanism implemented by 28
layer-18 MLP neurons. We use a complementary construction: each task circuit is
learned independently, the resulting masks share few neurons, and transfer is
tested between their activations.

Circuit analysis connects model behavior to internal components through
ablation, path patching, and sufficiency tests
\citep{elhage2021mathematical,wang2022interpretabilitywildcircuitindirect,conmy2023automatedcircuitdiscoverymechanistic}. Gradient-based
attribution can make this search cheaper, while causal evaluation remains
necessary because attribution alone does not establish that a component is
sufficient for the behavior~\citep{heimersheim2024useinterpretactivationpatching,kramar2024atpefficientscalablemethod}.
Sparse feature circuits extend this program to distributed representations
\citep{marks2025sparsefeaturecircuitsdiscovering}. Our masks operate on MLP neurons at a common layer so
that component overlap and activation transplantation are both defined
exactly.

Activation transplantation is closely related to interchange intervention.
In causal-abstraction work, an internal state from a source input replaces the
corresponding state in a base input; success tests whether the aligned neural
representation has the causal role assigned by a higher-level model
\citep{geiger2021causalabstractionsneuralnetworks,geiger2025causalabstractiontheoreticalfoundation}. Distributed alignment search
relaxes the requirement that high-level variables correspond to disjoint
neurons~\citep{pmlr-v236-geiger24a}. We align source and target examples by
normalized cycle phase and use the signed cyclic shift as the high-level operation. The
wrong-shift donor is therefore a direct counterfactual control: it preserves
the source task and intervention procedure while changing the proposed
computation.

Cross-task circuit reuse supplies the closest empirical comparison.
\citet{merullo2024circuitcomponentreusetasks} found 78\% attention-head overlap between circuits for
indirect-object identification and colored-object retrieval, then repaired one
task by editing four shared heads. That work establishes reuse through common
components. We test whether activations from independently discovered circuits
with little component overlap can causally repair behavior on another task
with analogous circular structure. The matched-versus-mismatched donor design
makes the test operation-specific.

\section{Experimental Design}
\label{sec:methods}

\subsection{Models and cyclic tasks}

We evaluate the instruction-tuned Qwen2.5-7B-Instruct and
Meta-Llama-3.1-8B-Instruct checkpoints without fine-tuning. The benchmark
contains alphabet positions, chess files, clock positions, compass bearings,
decimal digits, months, musical-note letters, seasons, and weekdays. A task
with cycle size $m$ asks for shifts $s\in\{-2,-1,+1,+2\}$. Each family has
three development templates and two held-out templates. Prompts state the
cycle order and sign convention and include balanced demonstrations.

All candidate answers are required to be unique one-token continuations. A
gold-token margin is the gold-token logit minus the largest logit among
incorrect task candidates. A
task enters the mechanistic analysis only when development and held-out
accuracy exceed chance by at least ten percentage points, every signed shift
has at least three correct development examples, at least two templates per
split contribute correct examples, and the held-out split contains at least
twelve correct examples. Later analyses use only correctly answered examples.

\subsection{Held-out circular geometry}

For a target index $y\in\{0,\ldots,m-1\}$, define its ideal circular coordinate
as
\begin{equation}
q(y)=\left(\cos\frac{2\pi y}{m},\;\sin\frac{2\pi y}{m}\right).
\end{equation}
At each residual-stream layer, we fit a ridge-regularized linear map from
centered hidden states to $q(y)$ using development templates. The layer is
selected on development data and frozen before evaluation. A task passes the
geometry gate only when both circular cosine and tangent consistency on the two
held-out templates exceed the 95th percentile of 100 label-shuffled fits.

\subsection{Sparse circuit discovery}

We score attention heads and MLP neurons with gradient times activation, then
learn sparse binary masks over post-gated MLP neurons at one shared comparison
layer: the median development-template attribution peak across qualified
tasks (zero-indexed layer 20 in Qwen and 14 in Llama). Each mask is optimized
for 64 steps with three random seeds and
deterministically pruned. We select the smallest mask whose median normalized
sufficiency is at least $0.80$ on selection prompts, then evaluate it on unseen
templates. If $M_{\mathrm{full}}$, $M_{\mathrm{only}}$, and
$M_{\mathrm{null}}$ denote the
gold-token logit margins of the full model, circuit-only model, and null model.
At the final prompt token, the circuit-only condition zeros unselected neurons
at this layer; the null condition zeros all neurons there. All other network
components remain active. Normalized sufficiency is
\begin{equation}
\frac{M_{\mathrm{only}}-M_{\mathrm{null}}}
     {M_{\mathrm{full}}-M_{\mathrm{null}}}.
\end{equation}
Twenty size- and activation-norm-matched random masks provide a nonspecific
circuit control. Circuit overlap is the conventional Jaccard similarity
between the independently selected neuron sets at this layer.

\subsection{Operators in circular coordinates}

For circuit mask $C$, let $z_C(x)$ be its selected post-gated MLP activations
and $W_{\mathrm{down},C}$ their output weights. Its residual contribution is
$\Delta h_C(x)=z_C(x)W_{\mathrm{down},C}^{\mathsf T}$. We project this contribution
into the task's frozen circular plane and fit an affine map
\begin{equation}
T_{C,s}(q)=A_{C,s}q+b_{C,s}
\end{equation}
from the input coordinate to the input coordinate plus the projected circuit
contribution. The fit uses one development template and is evaluated on held-out
templates. For cross-family comparisons, rotation angles are expressed in
cycle-step units before evaluating the operators on 64 common points. Our
continuous operator similarity is $\exp(-\operatorname{NRMSE})$ between the two
response curves, with no pairwise Procrustes realignment.

\subsection{Cross-task causal transfer}
\label{sec:transfer}

For each directed source-target task pair and signed shift, we run held-out
target prompts while zeroing the source-selected neurons at the shared layer
and final prompt token. Source examples are matched to the nearest circular
phase of the target's normalized input index $i/m$, and their selected MLP
activations replace those same neurons during the target run. The target mask
defines the neuron-overlap comparison. For an example with full, ablated,
and patched gold-token margins $M_f$, $M_a$, and $M_p$, the restoration score is
\begin{equation}
R=\frac{M_p-M_a}{M_f-M_a}.
\end{equation}
Examples with an ablation effect below $0.05$ are excluded, and each task-pair
and shift comparison requires at least four valid examples and a target
self-patch ceiling of at least
$0.80$. We compare the matched donor against three controls: phase-shuffled
source activations (a half-cycle phase offset), a donor from the same source
task performing the first available different shift in the order
$(-2,-1,+1,+2)$, and a size- and activation-norm-matched random mask.
Restoration scores are medians across valid examples within each task pair and
shift. Causal transfer specificity is the matched restoration score minus the
mean of these three controls; we average shifts within each directed pair,
then average across pairs.

The operation-selectivity contrast uses matched restoration minus wrong-shift
restoration. We first average shifts within each directed source-target task
pair, leaving 42 pair-level estimates for Qwen. Confidence
intervals resample these directed pairs. A random sign-flip test evaluates the
null hypothesis that the pair-level mean contrast is zero. Correlations between
operator similarity and transfer use the 161 eligible intervention rows and
are accompanied by task-pair bootstrap and shift-stratified permutation tests.

\section{Results}
\label{sec:results}

\subsection{Circular structure generalizes across prompts}

Qwen qualifies on seven of nine task families: alphabet, chess files, clock
positions, digits, months, musical notes, and weekdays (Figure~\ref{fig:funnel}).
Every qualified family
also passes both held-out geometry gates. Circular cosine ranges from $0.74$ to
$0.99$ (Figure~\ref{fig:geometry}). The fitted planes therefore recover the expected cyclic ordering from
prompt templates that were not used to select the layer or fit the plane.
All seven Qwen masks pass the prespecified sufficiency criterion on selection
prompts; three remain above $0.80$ on held-out prompts. Circuit identities are
frozen before held-out evaluation. The held-out score is reported only as a
robustness check; eligibility depends on selection sufficiency. Llama qualifies
on four families and passes held-out geometry on three.

\begin{figure}[t]
\floatconts
  {fig:funnel}
  {\caption{Task families retained at each analysis stage. Geometry is
  evaluated on templates excluded from layer and plane selection. Circuit
  sufficiency requires a median normalized score of at least $0.80$, first on
  selection prompts and then on held-out prompts.}}
  {\centering
  \begin{tikzpicture}[x=0.95cm,y=0.48cm,font=\scriptsize]
    % Two simple boxed plotting areas.
    \draw[black!55,line width=0.45pt] (0.9,0) rectangle (6.9,8.3);
    \draw[black!55,line width=0.45pt] (8.1,0) rectangle (14.1,8.3);

    % Shared vertical ticks, with labels only on the left panel.
    \foreach \y in {0,2,4,6,8} {
      \draw[black!55,line width=0.45pt] (0.9,\y) -- (1.03,\y);
      \draw[black!55,line width=0.45pt] (6.77,\y) -- (6.9,\y);
      \draw[black!55,line width=0.45pt] (8.1,\y) -- (8.23,\y);
      \draw[black!55,line width=0.45pt] (13.97,\y) -- (14.1,\y);
      \node[anchor=east,text=black!75] at (0.72,\y) {\y};
    }
    \node[rotate=90,font=\small] at (0.05,4.15) {Task families};

    % Meta-Llama bars.
    \filldraw[fill=barfill,draw=barstroke,line width=0.65pt]
      (1.35,0) rectangle (2.05,4);
    \filldraw[fill=barfill,draw=barstroke,line width=0.65pt]
      (2.75,0) rectangle (3.45,3);
    \filldraw[fill=barfill,draw=barstroke,line width=0.65pt]
      (4.15,0) rectangle (4.85,3);
    \node at (5.90,0.25) {0};
    \node[above] at (1.70,4) {4};
    \node[above] at (3.10,3) {3};
    \node[above] at (4.50,3) {3};

    % Qwen bars.
    \filldraw[fill=barfill,draw=barstroke,line width=0.65pt]
      (8.55,0) rectangle (9.25,7);
    \filldraw[fill=barfill,draw=barstroke,line width=0.65pt]
      (9.95,0) rectangle (10.65,7);
    \filldraw[fill=barfill,draw=barstroke,line width=0.65pt]
      (11.35,0) rectangle (12.05,7);
    \filldraw[fill=barfill,draw=barstroke,line width=0.65pt]
      (12.75,0) rectangle (13.45,3);
    \node[above] at (8.90,7) {7};
    \node[above] at (10.30,7) {7};
    \node[above] at (11.70,7) {7};
    \node[above] at (13.10,3) {3};

    % Panel titles and stage labels.
    \node[font=\small] at (3.9,9.0) {Meta-Llama 3.1 8B};
    \node[font=\small] at (11.1,9.0) {Qwen 2.5 7B};
    \foreach \x/\label in {
      1.70/Capability,
      3.10/Geometry,
      4.50/Selection,
      5.90/Held-out
    } {
      \node[rotate=27,anchor=north east] at (\x,-0.25) {\label};
    }
    \foreach \x/\label in {
      8.90/Capability,
      10.30/Geometry,
      11.70/Selection,
      13.10/Held-out
    } {
      \node[rotate=27,anchor=north east] at (\x,-0.25) {\label};
    }
  \end{tikzpicture}}
\end{figure}
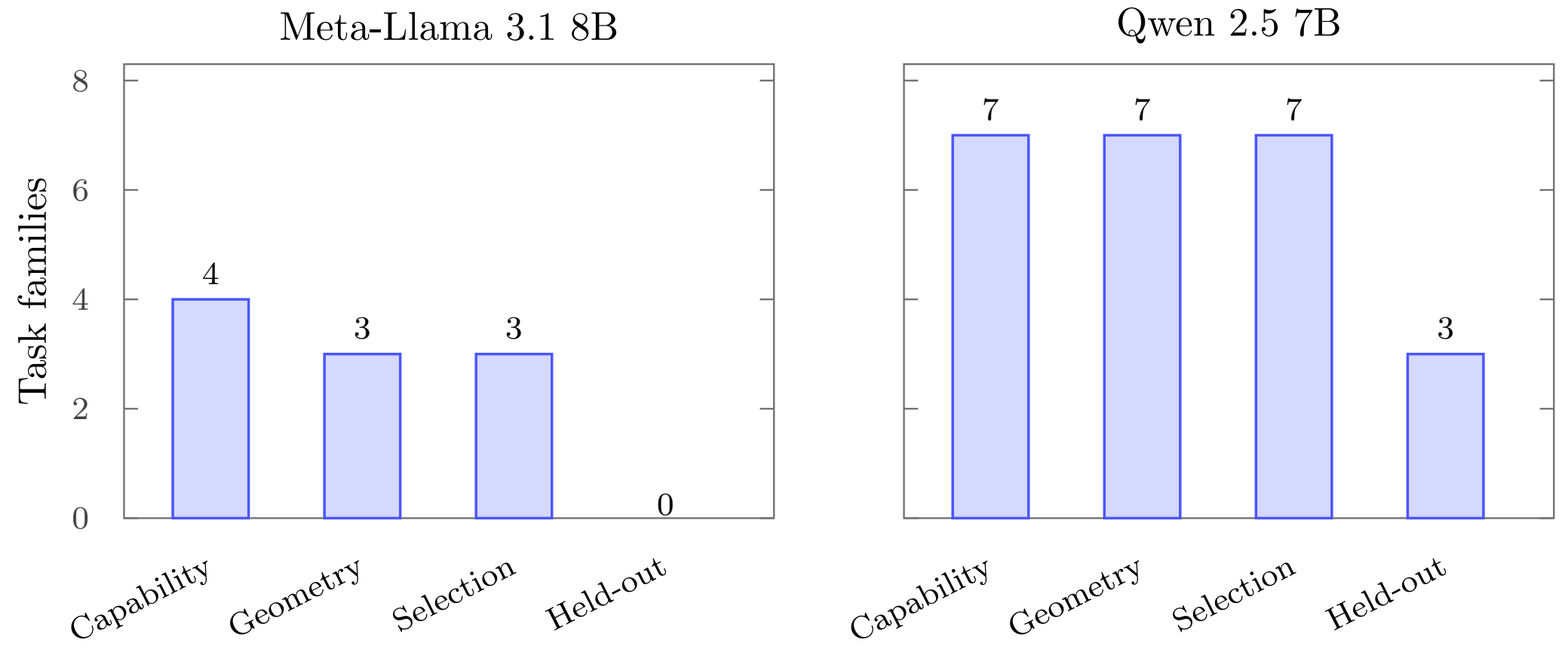

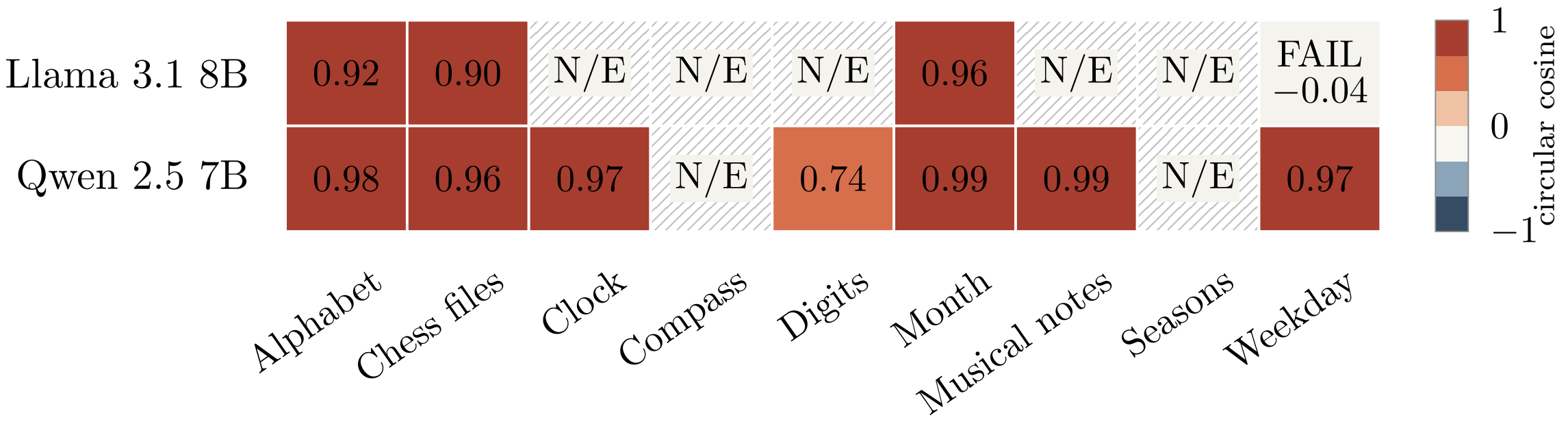
\begin{figure}[t]
\floatconts
  {fig:geometry}
  {\caption{Held-out circular geometry by task and model. Hatched cells did
  not meet the behavioral prerequisite. The labeled failure was evaluated but
  did not exceed both label-shuffled thresholds.}}
  {\centering
  \resizebox{\linewidth}{!}{%
  \begin{tikzpicture}[x=1.13cm,y=0.98cm,font=\small]
    \newcommand{\geomcell}[4]{%
      \path[fill=#3,draw=white,line width=0.7pt] (#1,#2) rectangle ++(1,1);
      \node at ($(#1,#2)+(0.5,0.5)$) {#4};}
    \newcommand{\missingcell}[2]{%
      \path[fill=heatmissing,draw=white,line width=0.7pt,
        pattern=north east lines,pattern color=black!24]
        (#1,#2) rectangle ++(1,1);
      \node[fill=heatmissing,inner sep=1.2pt] at ($(#1,#2)+(0.5,0.5)$) {N/E};}

    % Meta-Llama 3.1 8B (top row).
    \geomcell{0}{1}{heatposstrong}{0.92}
    \geomcell{1}{1}{heatposstrong}{0.90}
    \missingcell{2}{1}
    \missingcell{3}{1}
    \missingcell{4}{1}
    \geomcell{5}{1}{heatposstrong}{0.96}
    \missingcell{6}{1}
    \missingcell{7}{1}
    \geomcell{8}{1}{heatmissing}{\shortstack{FAIL\\$-0.04$}}

    % Qwen 2.5 7B (bottom row).
    \geomcell{0}{0}{heatposstrong}{0.98}
    \geomcell{1}{0}{heatposstrong}{0.96}
    \geomcell{2}{0}{heatposstrong}{0.97}
    \missingcell{3}{0}
    \geomcell{4}{0}{heatpos}{0.74}
    \geomcell{5}{0}{heatposstrong}{0.99}
    \geomcell{6}{0}{heatposstrong}{0.99}
    \missingcell{7}{0}
    \geomcell{8}{0}{heatposstrong}{0.97}

    \node[anchor=east,font=\normalsize] at (-0.18,1.5) {Llama 3.1 8B};
    \node[anchor=east,font=\normalsize] at (-0.18,0.5) {Qwen 2.5 7B};
    \foreach \x/\lab in {0/Alphabet,1/Chess files,2/Clock,3/Compass,
      4/Digits,5/Month,6/Musical notes,7/Seasons,8/Weekday} {
      \node[rotate=35,anchor=north east] at (\x+0.72,-0.12) {\lab};
    }

    % Compact continuous legend, separated from the data grid.
    \foreach \i/\c in {0/heatnegstrong,1/heatneg,2/heatzero,
      3/heatposlight,4/heatpos,5/heatposstrong} {
      \path[fill=\c,draw=none] (9.45,0.333*\i) rectangle (9.72,0.333*\i+0.34);
    }
    \draw[black!45,line width=0.35pt] (9.45,0) rectangle (9.72,2);
    \node[anchor=west] at (9.78,0) {$-1$};
    \node[anchor=west] at (9.78,1) {$0$};
    \node[anchor=west] at (9.78,2) {$1$};
    \node[rotate=90,font=\scriptsize] at (10.34,1) {circular cosine};
  \end{tikzpicture}}}
\end{figure}

\subsection{Low-overlap circuits support cross-task repair}

Qwen yields seven task families with valid primary circuits and 161 eligible
interventions across 42 directed source-target pairs. The aggregate transfer
estimates below use Qwen; Llama's ten eligible comparisons appear separately in
Figure~\ref{fig:transfer}. The task-specific masks share few neurons at the
comparison layer: median Jaccard similarity is $0.122$. Matched donors have
positive mean control-adjusted
transfer specificity at the task-pair level ($0.252$, 95\% bootstrap CI
$[0.162,0.349]$).

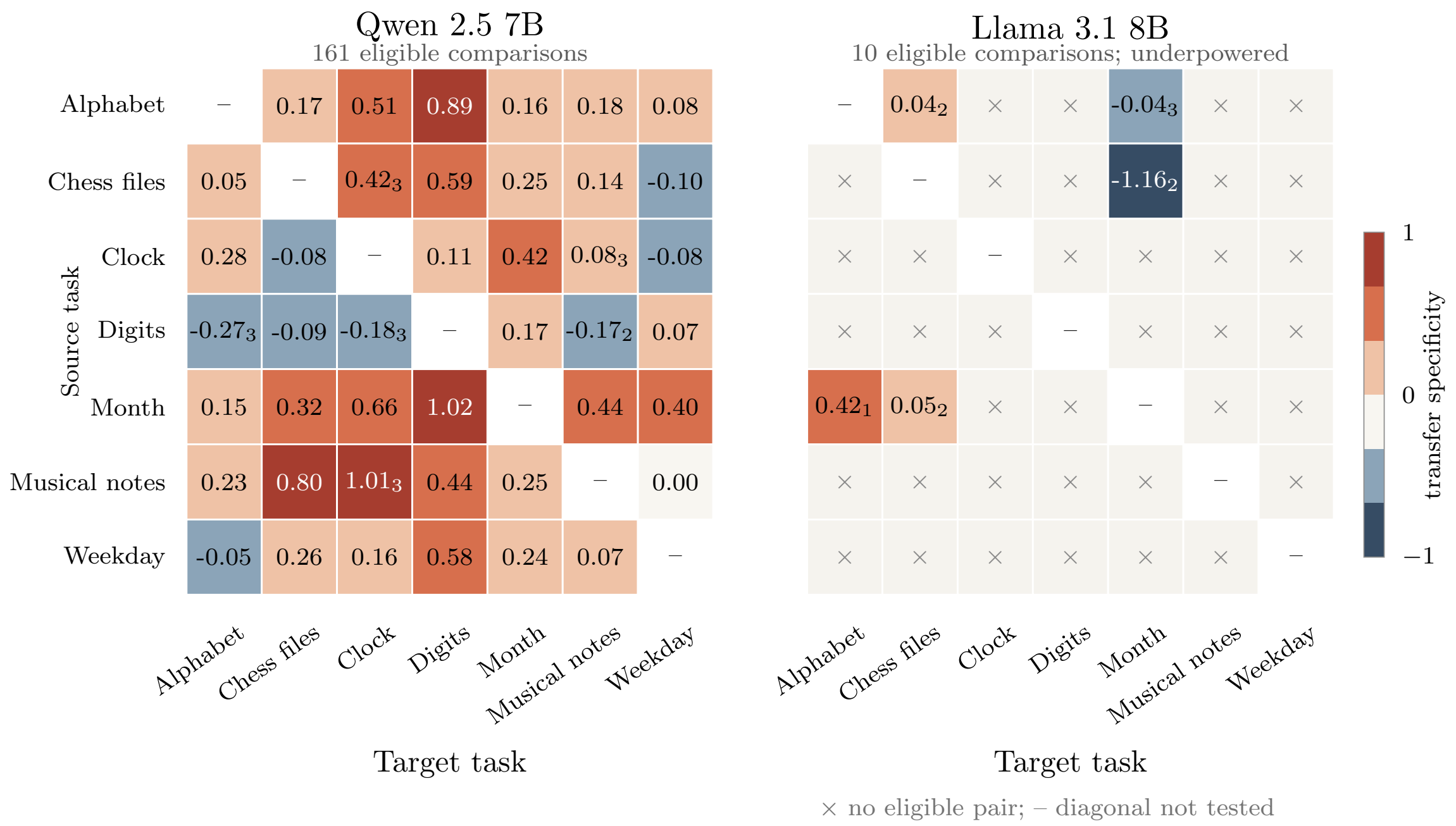
\begin{figure}[t]
\floatconts
  {fig:transfer}
  {\caption{Mean causal transfer specificity for each directed task pair.
  Positive values indicate that the matched source circuit repaired the
  target more effectively than the average of shuffled-donor, wrong-shift,
  and matched-random-mask controls. Cells average four shifts unless a
  subscript gives a smaller count.}}
  {\centering
  \resizebox{\linewidth}{!}{%
  \begin{tikzpicture}[x=0.86cm,y=0.86cm,font=\scriptsize]
    \newcommand{\tcell}[4]{%
      \path[fill=#3,draw=white,line width=0.6pt] (#1,#2) rectangle ++(1,1);
      \node[align=center] at ($(#1,#2)+(0.5,0.5)$) {#4};}
    \newcommand{\diagcell}[2]{\tcell{#1}{#2}{white}{--}}
    \newcommand{\emptycell}[2]{\tcell{#1}{#2}{heatmissing}{\textcolor{black!48}{$\times$}}}

    % Qwen 2.5 7B: rows are sources and columns are targets.
    \diagcell{0}{6} \tcell{1}{6}{heatposlight}{0.17}
    \tcell{2}{6}{heatpos}{0.51} \tcell{3}{6}{heatposstrong}{\textcolor{white}{0.89}}
    \tcell{4}{6}{heatposlight}{0.16} \tcell{5}{6}{heatposlight}{0.18}
    \tcell{6}{6}{heatposlight}{0.08}

    \tcell{0}{5}{heatposlight}{0.05} \diagcell{1}{5}
    \tcell{2}{5}{heatpos}{0.42$_3$} \tcell{3}{5}{heatpos}{0.59}
    \tcell{4}{5}{heatposlight}{0.25} \tcell{5}{5}{heatposlight}{0.14}
    \tcell{6}{5}{heatneg}{-0.10}

    \tcell{0}{4}{heatposlight}{0.28} \tcell{1}{4}{heatneg}{-0.08}
    \diagcell{2}{4} \tcell{3}{4}{heatposlight}{0.11}
    \tcell{4}{4}{heatpos}{0.42} \tcell{5}{4}{heatposlight}{0.08$_3$}
    \tcell{6}{4}{heatneg}{-0.08}

    \tcell{0}{3}{heatneg}{-0.27$_3$} \tcell{1}{3}{heatneg}{-0.09}
    \tcell{2}{3}{heatneg}{-0.18$_3$} \diagcell{3}{3}
    \tcell{4}{3}{heatposlight}{0.17} \tcell{5}{3}{heatneg}{-0.17$_2$}
    \tcell{6}{3}{heatposlight}{0.07}

    \tcell{0}{2}{heatposlight}{0.15} \tcell{1}{2}{heatpos}{0.32}
    \tcell{2}{2}{heatpos}{0.66} \tcell{3}{2}{heatposstrong}{\textcolor{white}{1.02}}
    \diagcell{4}{2} \tcell{5}{2}{heatpos}{0.44}
    \tcell{6}{2}{heatpos}{0.40}

    \tcell{0}{1}{heatposlight}{0.23} \tcell{1}{1}{heatposstrong}{\textcolor{white}{0.80}}
    \tcell{2}{1}{heatposstrong}{\textcolor{white}{1.01$_3$}} \tcell{3}{1}{heatpos}{0.44}
    \tcell{4}{1}{heatposlight}{0.25} \diagcell{5}{1}
    \tcell{6}{1}{heatzero}{0.00}

    \tcell{0}{0}{heatneg}{-0.05} \tcell{1}{0}{heatposlight}{0.26}
    \tcell{2}{0}{heatposlight}{0.16} \tcell{3}{0}{heatpos}{0.58}
    \tcell{4}{0}{heatposlight}{0.24} \tcell{5}{0}{heatposlight}{0.07}
    \diagcell{6}{0}

    % Meta-Llama 3.1 8B: sparse comparison panel.
    \begin{scope}[shift={(8.25,0)}]
      \diagcell{0}{6} \tcell{1}{6}{heatposlight}{0.04$_2$}
      \emptycell{2}{6}\emptycell{3}{6}\tcell{4}{6}{heatneg}{-0.04$_3$}
      \emptycell{5}{6}\emptycell{6}{6}
      \emptycell{0}{5}\diagcell{1}{5}\emptycell{2}{5}\emptycell{3}{5}
      \tcell{4}{5}{heatnegstrong}{\textcolor{white}{-1.16$_2$}}
      \emptycell{5}{5}\emptycell{6}{5}
      \emptycell{0}{4}\emptycell{1}{4}\diagcell{2}{4}\emptycell{3}{4}
      \emptycell{4}{4}\emptycell{5}{4}\emptycell{6}{4}
      \emptycell{0}{3}\emptycell{1}{3}\emptycell{2}{3}\diagcell{3}{3}
      \emptycell{4}{3}\emptycell{5}{3}\emptycell{6}{3}
      \tcell{0}{2}{heatpos}{0.42$_1$} \tcell{1}{2}{heatposlight}{0.05$_2$}
      \emptycell{2}{2}\emptycell{3}{2}\diagcell{4}{2}\emptycell{5}{2}\emptycell{6}{2}
      \emptycell{0}{1}\emptycell{1}{1}\emptycell{2}{1}\emptycell{3}{1}
      \emptycell{4}{1}\diagcell{5}{1}\emptycell{6}{1}
      \emptycell{0}{0}\emptycell{1}{0}\emptycell{2}{0}\emptycell{3}{0}
      \emptycell{4}{0}\emptycell{5}{0}\diagcell{6}{0}
    \end{scope}

    \foreach \y/\lab in {6/Alphabet,5/Chess files,4/Clock,3/Digits,
      2/Month,1/Musical notes,0/Weekday} {
      \node[anchor=east] at (-0.12,\y+0.5) {\lab};
    }
    \foreach \x/\lab in {0/Alphabet,1/Chess files,2/Clock,3/Digits,
      4/Month,5/Musical notes,6/Weekday} {
      \node[rotate=35,anchor=north east] at (\x+0.7,-0.14) {\lab};
      \node[rotate=35,anchor=north east] at (\x+8.95,-0.14) {\lab};
    }
    \node[rotate=90,font=\scriptsize] at (-1.55,3.5) {Source task};
    \node[font=\small] at (3.5,-2.25) {Target task};
    \node[font=\small] at (11.75,-2.25) {Target task};
    \node[font=\normalsize] at (3.5,7.55) {Qwen 2.5 7B};
    \node[font=\scriptsize,text=black!65] at (3.5,7.18) {161 eligible comparisons};
    \node[font=\normalsize] at (11.75,7.55) {Llama 3.1 8B};
    \node[font=\scriptsize,text=black!65] at (11.75,7.18) {10 eligible comparisons; underpowered};

    % Compact legend and availability key.
    \foreach \i/\c in {0/heatnegstrong,1/heatneg,2/heatzero,
      3/heatposlight,4/heatpos,5/heatposstrong} {
      \path[fill=\c,draw=none] (15.65,0.5+0.72*\i)
        rectangle (15.92,1.22+0.72*\i);
    }
    \draw[black!45,line width=0.35pt] (15.65,0.5) rectangle (15.92,4.82);
    \node[anchor=west] at (16.0,0.5) {$-1$};
    \node[anchor=west] at (16.0,2.66) {$0$};
    \node[anchor=west] at (16.0,4.82) {$1$};
    \node[rotate=90,font=\scriptsize] at (16.58,2.66) {transfer specificity};
    \node[anchor=west,text=black!55,font=\scriptsize] at (8.25,-2.85)
      {$\times$ no eligible pair; -- diagonal not tested};
  \end{tikzpicture}}}
\end{figure}

\subsection{Repair depends on matching the signed operation}

The operation-selectivity comparison changes the operation while holding the source task,
target task, intervention site, and patching procedure fixed. Correctly matched
donors restore $0.651$ more of the ablation-induced loss than wrong-shift donors
at the directed-task-pair level (95\% bootstrap CI $[0.474,0.838]$; sign-flip
$p<10^{-6}$). Both donors come from circuits acting on circular
representations. Their different outcomes estimate the effect of supplying the
matched rather than the selected wrong signed shift.

\subsection{Global operator similarity does not rank transfer strength}

The discrete matched-versus-wrong-shift contrast is strong, but the continuous
operator-similarity score is nearly uncorrelated with control-adjusted transfer
across Qwen's eligible rows (Spearman $\rho=0.031$). Operator similarity exceeds
circuit overlap as a transfer predictor by only $0.062$, with a task-pair
bootstrap interval spanning zero (95\% CI $[-0.239,0.369]$). The
shift-stratified comparison is also nonsignificant ($p=0.246$). Thus, the
discrete operation match captures a reliable causal contrast, while our finer
metric does not explain the remaining variation among task pairs. The row-level
comparison and the underpowered Llama result appear in
Appendix~\ref{app:operator-similarity}.

\section{Discussion}
\label{sec:discussion}

A circular manifold describes where concepts lie; a circuit supplies movement
between them. The matched- and wrong-shift donors differ in the operation they
supply. Their difference in repair therefore isolates the signed movement
carried by the transplanted activation more closely than a comparison against
shuffled or random activations could. This supports compatibility between
representation and computation.

Prior demonstrations of circuit reuse locate a common computation in shared
components. Here, source-circuit activations can support repair on target tasks
even when the independently learned source and target masks have little neuron
overlap and their representations are fit with separate circular planes.
Component identity and functional compatibility are consequently distinct
empirical questions: one asks which units are reused, while the other asks
whether a state produced by one circuit has the right causal effect on another
task.

The lack of association between continuous operator similarity and transfer
strength narrows the interpretation. The binary operation match captures a
strong causal preference in this benchmark, whereas small differences between the
fitted affine response curves do not reliably order the strength of repair.
This could reflect measurement error in the two-dimensional plane, coordinate
differences left by the frozen alignment, distributed computation outside the
MLP mask, or genuine target-specific compatibility not captured by the fitted
operator.

\section{Limitations}
\label{sec:limitations}

We study two instruction-tuned 7--8B models on nine synthetic cyclic task
families. Other architectures, scales, prompts, and reasoning problems may use
different representations or circuits. Qwen supplies most of the evidence;
Llama yields too few eligible comparisons for the main cross-pair analysis.

The circuit search is restricted to sparse MLP-neuron masks at selected
layers. Attention heads, edges, distributed mechanisms, or sparse features may
implement the same operation without appearing in these masks. Geometry is fit
only on correctly answered examples and summarized by a two-dimensional
Fourier plane. This isolates successful computation but can miss nonlinear
structure and information outside the fitted subspace.

The 161 Qwen intervention rows reuse circuits across shifts and task pairs.
They are not 161 independently discovered mechanisms. We therefore aggregate
the operation-selectivity result to 42 directed task pairs and preserve the
grouped structure in resampling tests. Finally, wrong-shift donors provide a
targeted operation control, but they can differ from matched donors in details
not captured by signed shift alone.

\section{Conclusion}
\label{sec:conclusion}

Circularly structured task representations support substantially stronger
cross-task repair when the transplanted activation carries a compatible
operation. This controlled intervention shows that organized circular
structure can remain present even when a transplanted activation supplies the
wrong transformation. Whether the same mismatch explains naturally occurring
errors remains open. Studies of representation geometry should therefore test
the stored state and the operation acting on it as separate causal objects.

\acks{We thank Professor H. Tad Blair of the University of California, Los
Angeles, for his invaluable feedback and thoughtful suggestions on this work.}

\renewcommand{\bibfont}{\small}
\bibliography{references}
\FloatBarrier
\clearpage

\appendix

\section{Continuous operator similarity}
\label{app:operator-similarity}

Figure~\ref{fig:operator-transfer} shows every eligible same-layer,
same-shift, cross-family intervention used in the correlation analysis. Qwen
provides 161 rows. Llama provides 10, below the prespecified minimum of 12 for
the cross-pair analysis.

\enlargethispage{3\baselineskip}
\noindent
\begin{minipage}{\linewidth}
  \centering
  \includegraphics[width=0.9\linewidth]{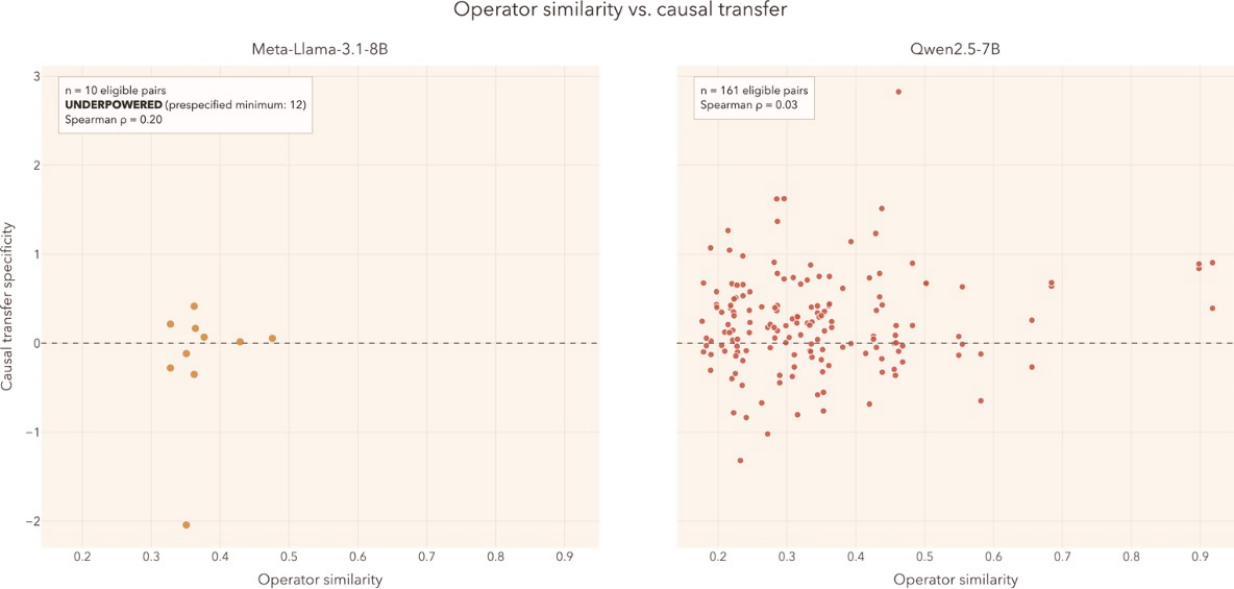}
  \captionof{figure}{Continuous operator similarity versus control-adjusted
  causal transfer. Qwen: row-level Spearman $\rho=0.031$; Llama: underpowered
  secondary comparison.}
  \label{fig:operator-transfer}
\end{minipage}

\end{document}